\documentclass[11pt,]{article}
\usepackage[left=1in,top=1in,right=1in,bottom=1in]{geometry}
\newcommand*{\authorfont}{\fontfamily{phv}\selectfont}
\usepackage[]{mathpazo}

  \usepackage[T1]{fontenc}
  \usepackage[utf8]{inputenc}

\usepackage{abstract}

\renewenvironment{abstract}
 {{%
    \setlength{\leftmargin}{0mm}
    \setlength{\rightmargin}{\leftmargin}%
  }%
  \relax}
 {\endlist}

\makeatletter
\def\@maketitle{%
  \newpage
  \let \footnote \thanks
    {\fontsize{18}{20}\selectfont\raggedright  \setlength{\parindent}{0pt} \@title \par}%
}
\makeatother

\usepackage{color}
\usepackage{fancyvrb}

\DefineVerbatimEnvironment{Highlighting}{Verbatim}{commandchars=\\\{\}}
\usepackage{framed}
\definecolor{shadecolor}{RGB}{248,248,248}
\newenvironment{Shaded}{\begin{snugshade}}{\end{snugshade}}

\newcommand{\CommentTok}[1]{\textcolor[rgb]{0.56,0.35,0.01}{\textit{#1}}}

\newcommand{\ControlFlowTok}[1]{\textcolor[rgb]{0.13,0.29,0.53}{\textbf{#1}}}
\newcommand{\DataTypeTok}[1]{\textcolor[rgb]{0.13,0.29,0.53}{#1}}
\newcommand{\DecValTok}[1]{\textcolor[rgb]{0.00,0.00,0.81}{#1}}

\newcommand{\ErrorTok}[1]{\textcolor[rgb]{0.64,0.00,0.00}{\textbf{#1}}}

\newcommand{\KeywordTok}[1]{\textcolor[rgb]{0.13,0.29,0.53}{\textbf{#1}}}
\newcommand{\NormalTok}[1]{#1}
\newcommand{\OperatorTok}[1]{\textcolor[rgb]{0.81,0.36,0.00}{\textbf{#1}}}
\newcommand{\OtherTok}[1]{\textcolor[rgb]{0.56,0.35,0.01}{#1}}

\newcommand{\StringTok}[1]{\textcolor[rgb]{0.31,0.60,0.02}{#1}}

\usepackage{longtable,booktabs}

\usepackage{graphicx,grffile}
\makeatletter
\def\maxwidth{\ifdim\Gin@nat@width>\linewidth\linewidth\else\Gin@nat@width\fi}
\def\maxheight{\ifdim\Gin@nat@height>\textheight\textheight\else\Gin@nat@height\fi}
\makeatother
\setkeys{Gin}{width=\maxwidth,height=\maxheight,keepaspectratio}

\title{On consistency scores in text data with an implementation in R \thanks{Thank you to Monica Alexander and Alex Luscombe for helpful comments. A Shiny application for interactive and small-scale examples is available: \url{https://kelichiu.shinyapps.io/Arianna/}. An R package for larger scale applications is available: \url{https://github.com/RohanAlexander/arianna}. Our code and datasets are available: \url{https://github.com/RohanAlexander/consistency_scores_in_text_datasets}. Comments on the 13 January 2021 version of this paper are welcome at: \href{mailto:rohan.alexander@utoronto.ca}{\nolinkurl{rohan.alexander@utoronto.ca}}.}  }

\author{\Large Ke-Li Chiu\vspace{0.05in} \newline\normalsize\emph{University of Toronto}   \and \Large Rohan Alexander\vspace{0.05in} \newline\normalsize\emph{University of Toronto}  }

\date{}

\usepackage{titlesec}

\usepackage{natbib}
\usepackage[strings]{underscore} 

\usepackage{setspace}

\makeatletter
\def\fps@figure{htbp}
\makeatother

\usepackage{hyperref}
\usepackage{amsmath}

\makeatletter
\@ifpackageloaded{hyperref}{}{%
\ifxetex
  \PassOptionsToPackage{hyphens}{url}\usepackage[setpagesize=false, 
              unicode=false, 
              xetex]{hyperref}
\else
  \PassOptionsToPackage{hyphens}{url}\usepackage[draft,unicode=true]{hyperref}
\fi
}

\@ifpackageloaded{color}{
    \PassOptionsToPackage{usenames,dvipsnames}{color}
}{%
    \usepackage[usenames,dvipsnames]{color}
}
\makeatother
\hypersetup{breaklinks=true,
            bookmarks=true,
            pdfauthor={Ke-Li Chiu (University of Toronto) and Rohan Alexander (University of Toronto)},
             pdfkeywords = {text-as-data; natural language processing; quantitative analysis; optical character recognition},  
            pdftitle={On consistency scores in text data with an implementation in R},
            colorlinks=true,
            citecolor=blue,
            urlcolor=blue,
            linkcolor=magenta,
            pdfborder={0 0 0}}
\usepackage[section]{placeins}

\begin{document}
	
%

{
\setlength{\parindent}{0pt}
\thispagestyle{plain}
{\fontsize{18}{20}\selectfont\raggedright 
\maketitle  

}

{
   \vskip 13.5pt\relax \normalsize\fontsize{11}{12} 
\textbf{\authorfont Ke-Li Chiu} \hskip 15pt \emph{\small University of Toronto}   \par \textbf{\authorfont Rohan Alexander} \hskip 15pt \emph{\small University of Toronto}   

}

}

\begin{abstract}

    \hbox{\vrule height .2pt width 39.14pc}

    \vskip 8.5pt 

\noindent In this paper, we introduce a reproducible cleaning process for the text extracted from PDFs using n-gram models. Our approach compares the originally extracted text with the text generated from, or expected by, these models using earlier text as stimulus. To guide this process, we introduce the notion of a consistency score, which refers to the proportion of text that is expected by the model. This is used to monitor changes during the cleaning process, and across different corpuses. We illustrate our process on text from the book Jane Eyre and introduce both a Shiny application and an R package to make our process easier for others to adopt.

 \par

    \hbox{\vrule height .2pt width 39.14pc}

\end{abstract}

\vskip -8.5pt


\noindent  

\hypertarget{introduction}{%
\section{Introduction}\label{introduction}}

When we think of quantitative analysis, we may like to think that our job is to `let the data speak'. But this is rarely the case in practise. Datasets can have errors, be biased, incomplete, or messy. In any case, it is the underlying statistical process, of which the dataset is an artifact, that is typically of interest. In order to use statistical models to understand that process, we typically need to clean and prepare the dataset in some way. This is especially the case when we work with text data because they are typically not designed for analysis. This cleaning and preparation requires us to make many decisions. To what extent should we correct obvious errors? What about slightly-less-obvious errors? Again, this is especially the case for text datasets because: they encode a large amount of information; an aspect may only be an error in context; and the high-dimensionality of the data at even its most basic level (26 English characters, compared with 10 numbers). Although cleaning and preparation is a necessary step, we may be concerned about the extent to which have we introduced new errors, and the possibility that we have made decisions that have affected, or even driven, our results.

In this paper we introduce the concept of consistency and implement it for a text corpus. Consistency refers to the proportion of data that are able to be forecast by a statistical model, based on preceding and surrounding data. Further, we define internal consistency as when the model is trained on the dataset itself, and external consistency as when the model is trained on a more general dataset. Together, these concepts provide a guide to the cleanliness and reasonableness of a dataset. This can be important when deciding whether a dataset is fit for purpose; when carrying out data cleaning and preparation tasks; and as a comparison between datasets. For instance, if the Canadian population is known to be 38 million from a census, but the sum of age-groups is initially 35 million, rising to 38 million only after an inputation error is corrected, then that post-cleaning dataset would have higher consistency than the pre-cleaning dataset. When applied to a text dataset, this means that we construct a model that forecasts the next word, given the previous words. If the forecast word is always the same as the actual word, then the consistency is 1. That said, it is not the absolute level of consistency that is our focus, but instead measuring how it changes during the cleaning and preparation process.

To provide an example, consider the sentence, `the cat in the\dots'. A child who has read this book could tell you that the next word should be `hat'. Hence if the sentence was actually `the cat in the bat', then that child would know something was likely wrong. The consistency score would likely be lower than if the sentence were `the cat in the hat' because we expected `hat', but actually found `bat'. After we correct this error, the consistency score would likely increase. By examining how consistency scores evolve in response to changes made to the text during the data preparation and cleaning stages we can better understand the effect of the changes. Including consistency scores and their evolution when datasets are shared allows researchers to be more transparent about the effect of their decisions in these steps of the analysis workflow. And finally, the use of consistency scores allows for an increased level of automation in the cleaning process.

We employ n-gram models to calculate consistency scores and generate word candidates for text corrections. The n-gram approach involves identifying words that are commonly found together. In our application, we use these sequences to calculate consistency scores and generate word candidates for text corrections. For instance, if we are using a sequence of three words, or tri-grams, then we identify the first two words in the sequence and evaluate if the last word in the sequence is `as expected' by comparing it to the corresponding tri-gram in a comparison dataset. The sequences we employ in this paper are a mixture of three, four, and five, consecutive words also known as tri-grams, 4-grams, and 5-grams.

The comparison dataset can be internal, meaning that the dataset is from the same source as the text being evaluated. On the other hand, a comparison dataset can be external, which means the dataset is from a different source, typically a larger and more universal dataset. The external dataset we use in the application discussed in this paper is constructed using a subset of text data sourced from Open WebText Corpus \citep{Gokaslan2019OpenWeb}. Since the content of Open WebText Corpus is from the internet, the variety of the text data is promising for meeting the generalizability and universality that an external dataset should possess, although it will also have extensive bias.

The resources that we have developed to support this paper include a Shiny app available at: \url{https://kelichiu.shinyapps.io/aRianna/}. That app computes internal and external consistency scores for corpus excerpts. We have also developed an R Package, \texttt{aRianna}, that allows our approach to be used on larger datasets, which is available at: \url{https://github.com/RohanAlexander/aRianna}.

In this paper, we introduce consistency scores and an R package that implements them. In Section \ref{background}, we provide an overview of language models and n-grams in particular. In Section \ref{arianna}, we deconstruct the package functions and provide demonstration example of the package usage. Finally, in Section \ref{discussion}, we discuss our findings, limitations, and future directions.

\hypertarget{background}{%
\section{Background}\label{background}}

\hypertarget{language-model}{%
\subsection{Language model}\label{language-model}}

Given the nature of human languages, some combinations of words tend to occur more frequently than others. Think of `good', which is more often followed by `morning', than `duck'. As such, we could consider English text production as a conditional probability, \(\mbox{Pr}(w_{k} | w^{k-1}_{1})\), where \(k\) is the number of words in a sentence, \(w_{k}\) is the predicted word, and \(w^{k-1}_{1}\) is the history of the word occurring in a sequence \citep{brown1992class}. In this way, the generation of some prediction, \(w_{k}\), is based on the history, \(w^{k-1}_{1}\). This is the underlying principle of all language models. Essentially, the goal of statistical language modeling is to estimate probability distributions over different linguistic units --- terms, words, sentences, and even documents \citep{bengio2003neural}. However, this is difficult as language is categorical. If we consider each word in a vocabulary as a category, then the dimensionality of a language dataset often becomes large quickly \citep{rosenfeld2000two}. One reason there is such a variety of statistical language models is that there are various ways of dealing with this fundamental problem. N-gram models are statistical language models that work by taking the co-occurrence of words in a sequence into account.

\hypertarget{n-gram-models}{%
\subsection{N-gram models}\label{n-gram-models}}

The foundation of an n-gram language model is the conditional probability set-up introduced above. An n-gram model is a probabilistic language model that predicts the next term in a sequence of terms \citep{bengio2003neural}. The \(n\) in n-gram refers to the number of terms in that sequence. Consider the following excerpt from \emph{Jane Eyre}: `We had been wandering'. `We' is a uni-gram, `We had' is a bi-gram, `We had been' is a tri-gram, and `We had been wandering' is a 4-gram. Notice that the two tri-grams in this excerpt, `We had been', and `had been wandering', overlap. The use of n-gram models enables us to assign probabilities to both the next sequence of words and just the next word. For instance, consider the two sentences: `We had been wandering', and `We had been wangling'. The former is likely to be more frequently encountered in a training corpus. Thus, a 4-gram would assign a higher probability to the next word being `wandering' than `wangling', given the sequence `We had been'.

To predict the next word, we have to take the sequence of preceding words into account, which requires knowing the probability of the sequence of words. The probability of a sequence appearing in a corpus follows the chain rule:
\begin{align*} 
\mbox{Pr(We, had, been, wandering)} = &  \mbox{Pr(We)} \times  \mbox{Pr(had|We)} \times  \mbox{Pr(been|We, had)} \\ 
 &  \times  \mbox{Pr(wandering|We, had, been)}.
\end{align*}

However, the likelihood that more and more words will occur next to each other in an identical sequence becomes smaller and smaller, making prediction difficult. Alternatively, we can approximate the probability of a word depending on only the previous word. This is known as the `Markov assumption' and it allows us to approximate the probability using only the last \(n\) words \citep{brown1992class}:
\[\mbox{Pr(We, had, been, wandering)} \approx \mbox{Pr(We)} \times \mbox{Pr(had|We)} \times \mbox{Pr(been|had)} \\ 
\times \mbox{Pr(wandering|been)}.\]

As a bi-gram model only considers the immediately preceding word, under the Markov assumption, an \(n\)-gram model can be reduced to a bi-gram model with \(n\) being any number:

\[\mbox{Pr}(w_{n} | w^{n-1}_{1}) \approx \mbox{Pr}(w_{n} | w_{n -1})\]

In our application, we use a model that has a mixture of tri-grams, 4-grams, and 5-grams to provide term replacement options for unexpected terms. Longer n-grams such as 5-grams capture more context and potentially generate more accurate predictions. However, there is less chance that a longer sequence of words will fit a specific combination which can lead to sparsity. Therefore, shorter n-grams should still be included to preserve the generalizability of the predictions.

Consider the idiom `elephant in the room'. If by mistake, this is written as `elephant in the roon' and correction is needed for the last word, then we can apply both a tri-gram model or a 4-gram model to predict the candidate word:

\begin{quote}
Using tri-gram model: `in the \texttt{\textless{}prediction\textgreater{}}'
\end{quote}

\begin{quote}
Using 4-gram model: `elephant in the \texttt{\textless{}prediction\textgreater{}}'
\end{quote}

In the 4-gram model, one more word, `elephant', is included, which provides more context and increased chance of `room' being accurately predicted. However, there are times when the literal location of an elephant may be being described; for instance, `elephant in the pond'. The combination of this sequence as a 4-gram might not be in a training dataset and no prediction would be available if only 4-gram models are used. Hence the use of tri-grams to provide other word candidates including `in the pond'.

\hypertarget{text-preparation-and-cleaning}{%
\subsection{Text preparation and cleaning}\label{text-preparation-and-cleaning}}

In general, text requires a great deal of preparation and cleaning before it can be analysed. For instance, \texttt{compositr} \citep{citecompositr} is an R package that runs through 13 common steps: remove non-UTF8 characters; insert missing comma spaces; replace misspelled words with most likely equivalent; remove HTML tags; convert case; remove URLs; remove numbers; remove punctuation; replace elongated words; remove extra white space; tokenization; stopword removal; and finally stem/lemmatize. And these are just some of the steps that may be taken.

Text preparation and cleaning steps and the uncertainty they bring, assume a fairly clean text dataset in the first instance. For instance, if a dataset has been created from Twitter then hashtags and emoji would need to be considered. And if the dataset had been created from OCR then there would likely be systematic errors to be addressed. One benefit of using packages such as \texttt{compositr} is that these decisions become repeatable and so their effect on subsequent analysis can be easily understood through repeated running of the full workflow. The notion of consistency provides a measure of this without the need to run the full workflow.

\hypertarget{consistency}{%
\subsection{Consistency}\label{consistency}}

Here we define, mathematically, what we mean by consistency in the case of text. Consider some text extract for which we want a consistency score. If there are \(n\) words, of which \(k\) are forecast correctly by a model, then the consistency score is the proportion: \(k/n\). We implement this in a dataframe using a binary column. Each word is a row, and a column, `correctly\_forecast', tracks the forecast, classifying each word as 1 if it was correctly forecast, and 0 otherwise. Then the internal consistency estimator is given by Equation \eqref{eq:model}:

\begin{equation} 
\label{eq:model}
c_i = \frac{\mbox{sum 'correctly_forecast' column}}{\mbox{length of dataset}}. 
\end{equation}

We distinguish between a model trained on the dataset for which consistency is being estimated, which results in an estimate of internal consistency, \(c_i\), and a model trained on another dataset, which results in an estimate of external consistency, \(c_e\), although the procedure is the same once the forecast has been made. We consider the forecast of each word independently, allowing for correlation in future work.

\hypertarget{arianna}{%
\section{\texorpdfstring{The \texttt{aRianna} R package}{The aRianna R package}}\label{arianna}}

\hypertarget{package-dependencies}{%
\subsection{Package dependencies}\label{package-dependencies}}

There are a variety of ways to implement an n-gram model within R \citep{citeR} including packages such as \texttt{quanteda} \citep{citequanteda}, \texttt{tidytext} \citep{citetidytext} and \texttt{tm} \citep{citetm}. We use \texttt{quanteda} because it has a comprehensive set of functions for conducting text analysis. We also use \texttt{tidyverse} \citep{citetidyverse} for data manipulation:

\begin{Shaded}
\begin{Highlighting}[]
\KeywordTok{install.packages}\NormalTok{(}\StringTok{"quanteda"}\NormalTok{)}
\KeywordTok{install.packages}\NormalTok{(}\StringTok{"tidyverse"}\NormalTok{)}
\KeywordTok{library}\NormalTok{(quanteda)}
\KeywordTok{library}\NormalTok{(tidyverse)}
\end{Highlighting}
\end{Shaded}

\hypertarget{function-to-make-the-internal-dataset}{%
\subsection{Function to make the internal dataset}\label{function-to-make-the-internal-dataset}}

To obtain the internal consistency score, we first construct a dataset from the text that is itself the text being evaluated. The function \texttt{aRianna::make\_internal\_consistency\_dataset()} is created for this purpose. The function turns a \texttt{body\_of\_text} into a dataset to compare the text with. To do so, the function first transforms the \texttt{body\_of\_text} into word tokens with the punctuation removed and the letters turned to lowercase:

\begin{Shaded}
\begin{Highlighting}[]
\NormalTok{make_internal_consistency_dataset <-}\StringTok{ }
\StringTok{  }\ControlFlowTok{function}\NormalTok{(body_of_text) \{}
\NormalTok{  tokens_from_example <-}\StringTok{ }
\StringTok{    }\NormalTok{quanteda}\OperatorTok{::}\KeywordTok{tokens}\NormalTok{(body_of_text, }\DataTypeTok{remove_punct =} \OtherTok{TRUE}\NormalTok{)}
\NormalTok{  tokens_from_example <-}\StringTok{ }
\StringTok{    }\NormalTok{quanteda}\OperatorTok{::}\KeywordTok{tokens_tolower}\NormalTok{(tokens_from_example)}
  \CommentTok{# code continues...}
\end{Highlighting}
\end{Shaded}

Next, the \texttt{aRianna::make\_internal\_consistency\_dataset()} function turns the individual tokens into tri-grams, 4-grams, and 5-grams, keeping only the n-grams that appear in the data for more than once. These common n-grams are stored in a tibble:

\begin{Shaded}
\begin{Highlighting}[]
  \CommentTok{# ...continue from previous}
  \CommentTok{# Create ngrams from the tokens}
\NormalTok{  toks_ngram <-}\StringTok{ }
\StringTok{    }\NormalTok{quanteda}\OperatorTok{::}\KeywordTok{tokens_ngrams}\NormalTok{(tokens_from_example, }\DataTypeTok{n =} \DecValTok{5}\OperatorTok{:}\DecValTok{3}\NormalTok{)}
  
  \CommentTok{# Convert to tibble so we can use our familiar verbs}
\NormalTok{  all_tokens <-}\StringTok{ }
\StringTok{    }\NormalTok{tibble}\OperatorTok{::}\KeywordTok{tibble}\NormalTok{(}\DataTypeTok{tokens =}\NormalTok{ toks_ngram[[}\DecValTok{1}\NormalTok{]])}
  
  \CommentTok{# We only want the common ones, not every one.}
\NormalTok{  all_tokens <-}
\StringTok{    }\NormalTok{all_tokens }\OperatorTok{%>%}
\StringTok{    }\NormalTok{dplyr}\OperatorTok{::}\KeywordTok{group_by}\NormalTok{(tokens) }\OperatorTok{%>%}
\StringTok{    }\NormalTok{dplyr}\OperatorTok{::}\KeywordTok{count}\NormalTok{() }\OperatorTok{%>%}
\StringTok{    }\NormalTok{dplyr}\OperatorTok{::}\KeywordTok{filter}\NormalTok{(n }\OperatorTok{>}\StringTok{ }\DecValTok{1}\NormalTok{) }\OperatorTok{%>%}
\StringTok{    }\NormalTok{dplyr}\OperatorTok{::}\KeywordTok{ungroup}\NormalTok{()}
  \CommentTok{# code continues...}
\end{Highlighting}
\end{Shaded}

Finally, the \texttt{aRianna::make\_internal\_consistency\_dataset()} function splits the n-grams into two parts: the first \(n-1\) words and the last word. As a result, a tibble is created with each n-gram containing its original sequence, the first words, and the last word:

\begin{Shaded}
\begin{Highlighting}[]
  \CommentTok{# ...continue from previous}
  \CommentTok{# Create a tibble that has the first two words in one column then the third}
\NormalTok{  all_tokens <-}
\StringTok{    }\NormalTok{all_tokens }\OperatorTok{%>%}
\StringTok{    }\NormalTok{dplyr}\OperatorTok{::}\KeywordTok{mutate}\NormalTok{(}\DataTypeTok{tokens =}\NormalTok{ stringr}\OperatorTok{::}\KeywordTok{str_replace_all}\NormalTok{(tokens, }\StringTok{"_"}\NormalTok{, }\StringTok{" "}\NormalTok{),}
                  \DataTypeTok{first_words =}\NormalTok{ stringr}\OperatorTok{::}\KeywordTok{word}\NormalTok{(tokens, }\DataTypeTok{start =} \DecValTok{1}\NormalTok{, }\DataTypeTok{end =} \DecValTok{-2}\NormalTok{),}
                  \DataTypeTok{last_word =}\NormalTok{ stringr}\OperatorTok{::}\KeywordTok{word}\NormalTok{(tokens, }\DecValTok{-1}\NormalTok{),}
                  \DataTypeTok{tokens =}\NormalTok{ stringr}\OperatorTok{::}\KeywordTok{str_replace_all}\NormalTok{(tokens, }\StringTok{" "}\NormalTok{, }\StringTok{"_"}\NormalTok{),}
                  \DataTypeTok{first_words =}\NormalTok{ stringr}\OperatorTok{::}\KeywordTok{str_replace_all}\NormalTok{(first_words, }\StringTok{" "}\NormalTok{, }\StringTok{"_"}\NormalTok{)}
\NormalTok{    ) }\OperatorTok{%>%}
\StringTok{    }\NormalTok{dplyr}\OperatorTok{::}\KeywordTok{rename}\NormalTok{(}\DataTypeTok{last_word_expected =}\NormalTok{ last_word) }\OperatorTok{%>%}
\StringTok{    }\NormalTok{dplyr}\OperatorTok{::}\KeywordTok{select}\NormalTok{(}\OperatorTok{-}\NormalTok{n)}
  \ErrorTok{\}}
\end{Highlighting}
\end{Shaded}

\hypertarget{function-to-generate-internal-consistency}{%
\subsection{Function to generate internal consistency}\label{function-to-generate-internal-consistency}}

After we have the internal consistency dataset, we need a function to compare the text being evaluated and the internal dataset in order to retrieve the internal consistency score. We define the \texttt{aRianna::generate\_internal\_consistency\_score} function to do this. Similar to the \texttt{aRianna::make\_internal\_consistency\_dataset} function, this function first transforms the text being evaluated into tri-grams, 4-grams, and 5-grams tokens, then splits the n-grams to their first words and last words:

\begin{Shaded}
\begin{Highlighting}[]
\NormalTok{generate_internal_consistency_score <-}
\StringTok{  }\ControlFlowTok{function}\NormalTok{(text_to_check, consistency_dataset)\{}
  \CommentTok{# Create tokens with errors}
\NormalTok{  tokens_from_example_with_errors <-}\StringTok{ }
\StringTok{    }\NormalTok{quanteda}\OperatorTok{::}\KeywordTok{tokens}\NormalTok{(text_to_check, }\DataTypeTok{remove_punct =} \OtherTok{TRUE}\NormalTok{)}
\NormalTok{  tokens_from_example_with_errors <-}\StringTok{ }
\StringTok{    }\NormalTok{quanteda}\OperatorTok{::}\KeywordTok{tokens_tolower}\NormalTok{(tokens_from_example_with_errors)}
  
  \CommentTok{# Create ngrams from the tokens with errors}
\NormalTok{  toks_ngram_with_errors <-}\StringTok{ }
\StringTok{    }\NormalTok{quanteda}\OperatorTok{::}\KeywordTok{tokens_ngrams}\NormalTok{(tokens_from_example_with_errors, }\DataTypeTok{n =} \DecValTok{5}\OperatorTok{:}\DecValTok{3}\NormalTok{)}
  
\NormalTok{  all_tokens_with_errors <-}\StringTok{ }
\StringTok{    }\NormalTok{tibble}\OperatorTok{::}\KeywordTok{tibble}\NormalTok{(}\DataTypeTok{tokens =}\NormalTok{ toks_ngram_with_errors[[}\DecValTok{1}\NormalTok{]])}
  
\NormalTok{  all_tokens_with_errors <-}
\StringTok{    }\NormalTok{all_tokens_with_errors }\OperatorTok{%>%}
\StringTok{    }\NormalTok{dplyr}\OperatorTok{::}\KeywordTok{mutate}\NormalTok{(}\DataTypeTok{ngram =} \KeywordTok{sapply}\NormalTok{(}\KeywordTok{strsplit}\NormalTok{(tokens, }\StringTok{"_"}\NormalTok{), length),}
                  \DataTypeTok{tokens =}\NormalTok{ stringr}\OperatorTok{::}\KeywordTok{str_replace_all}\NormalTok{(tokens, }\StringTok{"_"}\NormalTok{, }\StringTok{" "}\NormalTok{),}
                  \DataTypeTok{first_words =}\NormalTok{ stringr}\OperatorTok{::}\KeywordTok{word}\NormalTok{(tokens, }\DataTypeTok{start =} \DecValTok{1}\NormalTok{, }\DataTypeTok{end =} \DecValTok{-2}\NormalTok{),}
                  \DataTypeTok{last_word =}\NormalTok{ stringr}\OperatorTok{::}\KeywordTok{word}\NormalTok{(tokens, }\DecValTok{-1}\NormalTok{),}
                  \DataTypeTok{tokens =}\NormalTok{ stringr}\OperatorTok{::}\KeywordTok{str_replace_all}\NormalTok{(tokens, }\StringTok{" "}\NormalTok{, }\StringTok{"_"}\NormalTok{),}
                  \DataTypeTok{first_words =}\NormalTok{ stringr}\OperatorTok{::}\KeywordTok{str_replace_all}\NormalTok{(first_words, }\StringTok{" "}\NormalTok{, }\StringTok{"_"}\NormalTok{))}
  \CommentTok{# code continues...}
\end{Highlighting}
\end{Shaded}

Next, the function uses \texttt{dplyr::left\_join()} to combine the tibble of text tokens and the internal consistency tibble. The \texttt{dplyr::left\_join()} function returns all the rows from the text tokens tibble and all the columns from both tibbles \citep{2018dplyr}. By combining the two tibbles, the function generates two additional columns --- the \texttt{last\_word} column are the last words of all n-grams, and the \texttt{last\_word\_expected} contains the last words that are in the internal consistency dataset. By comparing the words in \texttt{last\_word} and \texttt{last\_word\_expected}, we can identify the `unexpected' words. If a last word is in \texttt{last\_word} but not \texttt{last\_word\_expected}, it is regarded as an unexpected word and the corresponding word in \texttt{last\_word\_expected} serves as a replacement candidate:

\begin{Shaded}
\begin{Highlighting}[]
  \CommentTok{# ...continue from previous}
\NormalTok{all_tokens_with_errors <-}
\StringTok{    }\NormalTok{all_tokens_with_errors }\OperatorTok{%>%}
\StringTok{    }\NormalTok{dplyr}\OperatorTok{::}\KeywordTok{left_join}\NormalTok{(dplyr}\OperatorTok{::}\KeywordTok{select}\NormalTok{(consistency_dataset, }\OperatorTok{-}\NormalTok{tokens), }
                     \DataTypeTok{by =} \KeywordTok{c}\NormalTok{(}\StringTok{"first_words"}\NormalTok{))}
  \CommentTok{# code continues...}
\end{Highlighting}
\end{Shaded}

The consistency score is calculated by the number of words that have been predicted by the model divided by the total number of words in the data. The function identifies the counts of words that are predicted by the model and divided by the total number of words in the input text:

\begin{Shaded}
\begin{Highlighting}[]
  \CommentTok{# ...continue from previous}
  \CommentTok{# Calculate the internal consistency score:}
\NormalTok{  false_count <-}\StringTok{ }
\StringTok{    }\KeywordTok{length}\NormalTok{(internal_consistency[}\KeywordTok{which}\NormalTok{(internal_consistency}\OperatorTok{$}\NormalTok{as_expected }\OperatorTok{==}\StringTok{ }\OtherTok{FALSE} \OperatorTok{&}\StringTok{ }
\StringTok{                                        }\NormalTok{internal_consistency}\OperatorTok{$}\NormalTok{ngram }\OperatorTok{==}\StringTok{ }\DecValTok{3}\NormalTok{)])}
\NormalTok{  word_count <-}\StringTok{ }\KeywordTok{sapply}\NormalTok{(}\KeywordTok{strsplit}\NormalTok{(text_to_check, }\StringTok{" "}\NormalTok{), length)}
\NormalTok{  true_count <-}\StringTok{ }\NormalTok{word_count }\OperatorTok{-}\StringTok{ }\NormalTok{false_count}

\NormalTok{  internal_consistency <-}\StringTok{ }\NormalTok{tibble}\OperatorTok{::}\KeywordTok{tibble}\NormalTok{(}
    \StringTok{"as_expected"}\NormalTok{ =}\StringTok{ }\NormalTok{true_count,}
    \StringTok{"unexpected"}\NormalTok{  =}\StringTok{ }\NormalTok{false_count,}
    \StringTok{"consistency"}\NormalTok{ =}\StringTok{ }\NormalTok{true_count}\OperatorTok{/}\NormalTok{word_count}
\NormalTok{  )}
  \CommentTok{# code continues...}
\end{Highlighting}
\end{Shaded}

The \texttt{aRianna::generate\_internal\_consistency\_score} function also lists the identified text errors and generates replacement candidates to the text errors:

\begin{Shaded}
\begin{Highlighting}[]
  \CommentTok{# ...continue from previous}
  \CommentTok{# Identify which words were unexpected}
\NormalTok{  unexpected <-}
\StringTok{    }\NormalTok{all_tokens_with_errors_only }\OperatorTok{%>%}
\StringTok{    }\NormalTok{dplyr}\OperatorTok{::}\KeywordTok{mutate}\NormalTok{(}\DataTypeTok{as_expected =}\NormalTok{ last_word }\OperatorTok{==}\StringTok{ }\NormalTok{last_word_expected) }\OperatorTok{%>%}
\StringTok{    }\NormalTok{dplyr}\OperatorTok{::}\KeywordTok{filter}\NormalTok{(as_expected }\OperatorTok{==}\StringTok{ }\OtherTok{FALSE}\NormalTok{) }\OperatorTok{%>%}
\StringTok{    }\NormalTok{dplyr}\OperatorTok{::}\KeywordTok{select}\NormalTok{(}\OperatorTok{-}\NormalTok{tokens, }\OperatorTok{-}\NormalTok{ngram, }\OperatorTok{-}\NormalTok{as_expected)}
  \ErrorTok{\}}
\end{Highlighting}
\end{Shaded}

\hypertarget{function-to-generate-external-consistency}{%
\subsection{Function to generate external consistency}\label{function-to-generate-external-consistency}}

The \texttt{aRianna::generate\_external\_consistency\_score} function works identically. The only difference is that it compares the text being evaluated with an external consistency dataset that is larger and more general. The external consistency dataset we employed in the application is constructed using a subset of text data from Open WebText Corpus \citep{Gokaslan2019OpenWeb}. In general, any dataset could be used.

\hypertarget{demonstration}{%
\subsection{Demonstration}\label{demonstration}}

To install the package and load the library:

\begin{Shaded}
\begin{Highlighting}[]
\NormalTok{devtools}\OperatorTok{::}\KeywordTok{install_github}\NormalTok{(}\StringTok{"RohanAlexander/arianna"}\NormalTok{)}
\KeywordTok{library}\NormalTok{(aRianna)}
\end{Highlighting}
\end{Shaded}

For this demonstration, we use the first paragraph of Jane Eyre as the internal text data. A sentence within the paragraph is modified with the intention to contain an error and serves as the text to be evaluated:

\begin{Shaded}
\begin{Highlighting}[]
\NormalTok{body_of_text <-}\StringTok{ "There was no possibility of taking a walk that day. }
\StringTok{  We had been wandering, indeed, in the leafless shrubbery an hour in }
\StringTok{  the morning; but since dinner (Mrs. Reed, when there was no company, }
\StringTok{  dined early) the cold winter wind had brought with it clouds so sombre, }
\StringTok{  and a rain so penetrating, that further out-door exercise was now out }
\StringTok{  of the question."}

\NormalTok{text_to_evaluate <-}\StringTok{ "when there was na company"}
\end{Highlighting}
\end{Shaded}

The first step is to turn the body of text into the internal consistency dataset. The generated internal consistency dataset is a tibble that contains the tri-grams, 4-grams and 5-grams that appear in the internal text data more than once. Since only `there\_was\_no' has more than one occurrence, it is the only n-gram in the internal consistency dataset:

\begin{Shaded}
\begin{Highlighting}[]
\NormalTok{internal_consistency_dataset <-}\StringTok{ }
\StringTok{  }\NormalTok{aRianna}\OperatorTok{::}\KeywordTok{make_internal_consistency_dataset}\NormalTok{(body_of_text)}
\NormalTok{internal_consistency_dataset}
\CommentTok{## # A tibble: 1 x 3}
\CommentTok{##   tokens       first_words last_word_expected}
\CommentTok{##   <chr>        <chr>       <chr>             }
\CommentTok{## 1 there_was_no there_was   no                }
\end{Highlighting}
\end{Shaded}

The next step is to compare the text to be evaluated with the internal consistency dataset that we created in the previous step. The function \texttt{aRianna::generate\_internal\_consistency\_score()} takes two arguments: the text to evaluate, and the internal consistency dataset. The function identifies the word `na' as an unexpected word and generates the internal consistency score as 0.8. The function also lists `no' as the replacement of `na':

\begin{Shaded}
\begin{Highlighting}[]
\NormalTok{aRianna}\OperatorTok{::}\KeywordTok{generate_internal_consistency_score}\NormalTok{(}
\NormalTok{  text_to_evaluate, internal_consistency_dataset)}

\CommentTok{## $`internal consistency`}
\CommentTok{##   as_expected unexpected consistency}
\CommentTok{## 1           4          1         0.8}

\CommentTok{## $`unexpected words`}
\CommentTok{## # A tibble: 1 x 3}
\CommentTok{##   first_words last_word last_word_expected }
\CommentTok{##   <chr>       <chr>     <chr>                 }
\CommentTok{## 1 there_was   na        no                 }
\end{Highlighting}
\end{Shaded}

To get the external consistency score, we use the \texttt{aRianna::generate\_external\_consistency\_score()} function. The function takes only one argument, the text to evaluate, and compares it with the external consistency dataset. Here, the text to evaluate is `there was no possibiliti', with the word ``possibility'' being wrongly spelled. The function identifies the word `possibiliti' as an unexpected word, and generates the external consistency score 0.75. Because the external dataset is larger, it is capable of providing more replacement candidates for `possibiliti'. There are no replacements based on 5-grams, so the replacements based on 4-grams are displayed first, followed by the replacements based on tri-grams:

\begin{Shaded}
\begin{Highlighting}[]
\NormalTok{text_to_evaluate <-}\StringTok{ "there was no possibiliti"}
\NormalTok{aRianna}\OperatorTok{::}\KeywordTok{generate_external_consistency_score}\NormalTok{(text_to_evaluate)}

\CommentTok{## aRianna::generate_external_consistency_score(text_to_evaluate)}
\CommentTok{## $`external consistency`}
\CommentTok{## # A tibble: 1 x 3}
\CommentTok{##   as_expected unexpected consistency}
\CommentTok{##         <int>      <int>       <dbl>}
\CommentTok{## 1           3          1        0.75}

\CommentTok{## $`unexpected words`}
\CommentTok{## # A tibble: 11 x 3}
\CommentTok{##    first_words  last_word    last_word_expected }
\CommentTok{##    <chr>        <chr>        <chr>                    }
\CommentTok{##  1 there_was_no possibiliti  evidence          }
\CommentTok{##  2 there_was_no possibiliti  immediate         }
\CommentTok{##  3 there_was_no possibiliti  infrastructure    }
\CommentTok{##  4 there_was_no possibiliti  one               }
\CommentTok{##  5 there_was_no possibiliti  possibility       }
\CommentTok{##  6 there_was_no possibiliti  sound             }
\CommentTok{##  7 there_was_no possibiliti  way               }
\CommentTok{##  8 was_no       possibiliti  good              }
\CommentTok{##  9 was_no       possibiliti  longer            }
\CommentTok{## 10 was_no       possibiliti  more              }
\CommentTok{## 11 was_no       possibiliti  wonder }
\end{Highlighting}
\end{Shaded}

The choice of what do now depends on a variety of factors including the purpose of the analysis, and the size of the corpus. One option is to consider only a subset of potential replacements, or to consider only replacements over a certain probability threshold. These are parameters that can be tuned in more sophisticated language models. The choice of which word to change `possibili' to could affect the consistency score if it is not what the model expects. Future work, with more sophisticated language models, could allow tuning in this regard.

As the dataset increases in size, the number of possible replacements will also likely increase. Various options for decreasing the number of possible replacements exist, starting with string distance algorithms and increasing in complexity. It is important to recognise that we are focused on the fact that they were not forecast, rather than what they should be replaced with. However, improving this aspect helps to make the consistency score more relevant and useful, especially in text settings where the cleaning and preparation aspect is likely to be iterative.

\hypertarget{discussion}{%
\section{Discussion}\label{discussion}}

Outside of textbook examples, almost all real-world datasets need to be cleaned and prepared before they can be analyzed. While there are many tools to help with reproducibility of analysis, options for ensuring reproducibility in earlier stages of a typical data science workflow are more limited. Our approach discussed in this paper can help. Our notion of consistency can be tracked at many points while the dataset is being cleaned and prepared. Stages in which it changes substantially can be investigated. While there may be good reasons for such as change, by identifying them it is easier for other researchers to understand the effects of cleaning and preparation on eventual analysis results.

Our approaches are applicable to any dataset, but in this paper we applied them to one based on text. The high-dimensionality of text, and the messiness of it when it comes from real-world sources, means that often considerable cleaning and preparation is done to text datasets. We demonstrated how sequences of words can be used to forecast the very next word. That forecast can be compared with the actual word observed and this difference expressed as a proportion, that we call a `consistency score'.

The language model that we used is n-grams. Language models underpinned by n-grams are widely applied in text prediction, spelling-correction, and machine translation \citep{brown1992class}. As demonstrated in our application, we use a n-gram based model to calculate internal and external consistency scores and generate a list of words as text correction candidates. However, n-gram models do not take the linguistic structure of language, nor broader context, into account. For instance, \citet[p.~1]{rosenfeld2000two} discusses language in this context, saying that `\ldots it may as well be a sequence of arbitrary symbols, with no deep structure, intention or thought behind'. Hence next-word prediction using n-gram-based language models can be limited. Here think of a two-gram involving the word `good' such as `good morning'. At scale, these can identify missing or unusual words, and work quickly, but they lack nuance. For instance, an equally reasonable two-gram involving the word `good' is `good work'. However, because n-gram models do not take the surrounding text into account, they do not have the capacity to judge if `morning' or `work' should be suggested as the next word after `good'.

In future work we intend to investigate more advanced language models such as word embeddings and Transformers. We will consider pre-trained word embedding models including Word2Vec \citep{mikolov2013efficient} \citep{mikolov2013distributed} and GloVe \citep{pennington2014glove}, which place each word in a multi-dimensional space such that distance between words can illustrate their relationship \citep{bengio2003neural}. This approach of representing words in vectors has a long history, but the implementation of \citet{bengio2003neural} has become the foundation for much subsequent work. For instance, \citet{stoltz2019concept} use word embeddings to identify how documents relate to concepts.

We will also consider incorporating generative pre-trained Transformer models such as GPT-2 \citep{radford2019language}, GPT-3 \citep{brown2020language}, and BERT \citep{devlin2018bert}. These models are based on the Transformer architecture \citep{vaswani2017attention}. Proposed by \citet{vaswani2017attention} in 2017, the Transformer network architecture for neural networks has been found to outperform both recurrent neural network-based and convolutional neural network-based models in computational efficiency \citep{vaswani2017attention}. Since the emergence of the Transformer model, most of the representative pre-trained models are built on this architecture \citep{hanretty2018comparing}.

Although the level of the consistency score itself is not of major importance compared with its changes, improving the score by using more sophisticated models would be beneficial in terms of identifying incorrect words. In our future work we are interested in implementing different language models to compare their effectiveness in the task of generating consistency scores and providing suggested text corrections. We intend to use our future findings to continuously improve the \texttt{aRianna} package. We are also interested in expanding the notion of consistency beyond text and applying it to other datasets.

\newpage

\newpage
\singlespacing 
\renewcommand\refname{References}
\bibliography{references.bib}

\begin{thebibliography}{}

\bibitem[Bengio et~al., 2003]{bengio2003neural}
Bengio, Y., Ducharme, R., Vincent, P., and Jauvin, C. (2003).
\newblock A neural probabilistic language model.
\newblock {\em {Journal of Machine Learning Research}}, 3(Feb):1137--1155.

\bibitem[Benoit et~al., 2018]{citequanteda}
Benoit, K., Watanabe, K., Wang, H., Nulty, P., Obeng, A., Müller, S., and
  Matsuo, A. (2018).
\newblock quanteda: An r package for the quantitative analysis of textual data.
\newblock {\em Journal of Open Source Software}, 3(30):774.

\bibitem[Brown et~al., 1992]{brown1992class}
Brown, P.~F., Della~Pietra, V.~J., Desouza, P.~V., Lai, J.~C., and Mercer,
  R.~L. (1992).
\newblock Class-based n-gram models of natural language.
\newblock {\em Computational linguistics}, 18(4):467--480.

\bibitem[Brown et~al., 2020]{brown2020language}
Brown, T.~B., Mann, B., Ryder, N., Subbiah, M., Kaplan, J., Dhariwal, P.,
  Neelakantan, A., Shyam, P., Sastry, G., Askell, A., et~al. (2020).
\newblock Language models are few-shot learners.
\newblock {\em arXiv preprint arXiv:2005.14165}.

\bibitem[Devlin et~al., 2018]{devlin2018bert}
Devlin, J., Chang, M.-W., Lee, K., and Toutanova, K. (2018).
\newblock Bert: Pre-training of deep bidirectional transformers for language
  understanding.
\newblock {\em arXiv preprint arXiv:1810.04805}.

\bibitem[Feinerer and Hornik, 2019]{citetm}
Feinerer, I. and Hornik, K. (2019).
\newblock {\em tm: Text Mining Package}.
\newblock R package version 0.7-7.

\bibitem[Gokaslan and Cohen, 2019]{Gokaslan2019OpenWeb}
Gokaslan, A. and Cohen, V. (2019).
\newblock Openwebtext corpus.
\newblock \url{http://Skylion007.github.io/OpenWebTextCorpus}.

\bibitem[Hanretty et~al., 2018]{hanretty2018comparing}
Hanretty, C., Lauderdale, B.~E., and Vivyan, N. (2018).
\newblock Comparing strategies for estimating constituency opinion from
  national survey samples.
\newblock {\em Political Science Research and Methods}, 6(3):571--591.

\bibitem[Luscombe, 2021]{citecompositr}
Luscombe, A. (2021).
\newblock {\em compositr: Efficient tools for preprocessing text data in R.}
\newblock R package version 0.0.0.9000.

\bibitem[Mikolov et~al., 2013a]{mikolov2013efficient}
Mikolov, T., Chen, K., Corrado, G., and Dean, J. (2013a).
\newblock Efficient estimation of word representations in vector space.
\newblock {\em arXiv preprint arXiv:1301.3781}.

\bibitem[Mikolov et~al., 2013b]{mikolov2013distributed}
Mikolov, T., Sutskever, I., Chen, K., Corrado, G.~S., and Dean, J. (2013b).
\newblock Distributed representations of words and phrases and their
  compositionality.
\newblock In {\em Advances in neural information processing systems}, pages
  3111--3119.

\bibitem[Pennington et~al., 2014]{pennington2014glove}
Pennington, J., Socher, R., and Manning, C.~D. (2014).
\newblock Glove: Global vectors for word representation.
\newblock In {\em Proceedings of the 2014 conference on empirical methods in
  natural language processing (EMNLP)}, pages 1532--1543.

\bibitem[{R Core Team}, 2019]{citeR}
{R Core Team} (2019).
\newblock {\em R: A Language and Environment for Statistical Computing}.
\newblock R Foundation for Statistical Computing, Vienna, Austria.

\bibitem[Radford et~al., 2019]{radford2019language}
Radford, A., Wu, J., Child, R., Luan, D., Amodei, D., and Sutskever, I. (2019).
\newblock Language models are unsupervised multitask learners.
\newblock {\em OpenAI Blog}, 1(8):9.

\bibitem[Rosenfeld, 2000]{rosenfeld2000two}
Rosenfeld, R. (2000).
\newblock Two decades of statistical language modeling: Where do we go from
  here?
\newblock {\em Proceedings of the IEEE}, 88(8):1270--1278.

\bibitem[Silge and Robinson, 2016]{citetidytext}
Silge, J. and Robinson, D. (2016).
\newblock tidytext: Text mining and analysis using tidy data principles in r.
\newblock {\em JOSS}, 1(3).

\bibitem[Stoltz and Taylor, 2019]{stoltz2019concept}
Stoltz, D.~S. and Taylor, M.~A. (2019).
\newblock Concept mover’s distance: measuring concept engagement via word
  embeddings in texts.
\newblock {\em Journal of Computational Social Science}, 2(2):293--313.

\bibitem[Vaswani et~al., 2017]{vaswani2017attention}
Vaswani, A., Shazeer, N., Parmar, N., Uszkoreit, J., Jones, L., Gomez, A.~N.,
  Kaiser, {\L}., and Polosukhin, I. (2017).
\newblock Attention is all you need.
\newblock In {\em Advances in neural information processing systems}, pages
  5998--6008.

\bibitem[Wickham et~al., 2019]{citetidyverse}
Wickham, H., Averick, M., Bryan, J., Chang, W., McGowan, L.~D., François, R.,
  Grolemund, G., Hayes, A., Henry, L., Hester, J., Kuhn, M., Pedersen, T.~L.,
  Miller, E., Bache, S.~M., Müller, K., Ooms, J., Robinson, D., Seidel, D.~P.,
  Spinu, V., Takahashi, K., Vaughan, D., Wilke, C., Woo, K., and Yutani, H.
  (2019).
\newblock Welcome to the {tidyverse}.
\newblock {\em Journal of Open Source Software}, 4(43):1686.

\bibitem[Wickham et~al., 2018]{2018dplyr}
Wickham, H., François, R., Henry, L., and Müller, K. (2018).
\newblock {\em dplyr: A Grammar of Data Manipulation}.
\newblock R package version 0.7.6.

\end{thebibliography}

\end{document}